\documentclass{article}

\usepackage[preprint]{neurips_2024}

\usepackage[utf8]{inputenc} 
\usepackage[T1]{fontenc}    
\usepackage{url}            
\usepackage{booktabs}       
\usepackage{amsfonts}       
\usepackage{nicefrac}       
\usepackage{microtype}      
\usepackage{xcolor}         
\usepackage{graphicx}
\usepackage{amsmath}
\usepackage{adjustbox}
\usepackage{subcaption}
\usepackage{pifont} 
\usepackage{tikz} 
\usepackage{siunitx} 
\usepackage{wrapfig} 
\usepackage{enumitem}
\usepackage{multirow}
\usepackage{natbib}
\setcitestyle{numbers,square}

\definecolor{lightblue}{RGB}{54, 116, 230}
\usepackage[colorlinks, linkcolor=red, citecolor=lightblue]{hyperref}

\title{UCDNet: Multi-UAV Collaborative 3D Object Detection Network by Reliable Feature Mapping}

\author{
	Pengju Tian$^{1,2}$, 
	Peirui Cheng$^{1}$, 
	Yuchao Wang$^{1,2}$, 
	Zhechao Wang$^{1,2}$, 
	Zhirui Wang$^{1}$\thanks{Corresponding author} \\
	\textbf{Menglong Yan}$^{1}$, 
	\textbf{Xue Yang}$^{3}$, 
	\textbf{Xian Sun}$^{1,2}$
	\\
	$^1$Key Laboratory of Network Information System Technology,
	\\ Aerospace Information Research Institute, Chinese Academy of Sciences \\
        $^2$University of Chinese Academy of Sciences \quad $^3$Shanghai AI Laboratory \\
	\texttt{\small tianpengju22@mails.ucas.ac.cn}\\
 }

\begin{document}

\maketitle

\begin{abstract}
Multi-UAV collaborative 3D object detection can perceive and comprehend complex environments by integrating complementary information, with applications encompassing traffic monitoring, delivery services and agricultural management. However, the extremely broad observations in aerial remote sensing and significant perspective differences across multiple UAVs make it challenging to achieve precise and consistent feature mapping from 2D images to 3D space in multi-UAV collaborative 3D object detection paradigm. To address the problem, we propose an unparalleled camera-based multi-UAV collaborative 3D object detection paradigm called UCDNet. Specifically, the depth information from the UAVs to the ground is explicitly utilized as a strong prior to provide a reference for more accurate and generalizable feature mapping. Additionally, we design a homologous points geometric consistency loss as an auxiliary self-supervision, which directly influences the feature mapping module, thereby strengthening the global consistency of multi-view perception. Experiments on AeroCollab3D and CoPerception-UAVs datasets show our method increases 4.7$\%$ and 10$\%$ mAP respectively compared to the baseline, which demonstrates the superiority of UCDNet.

\end{abstract}

\section{Introduction}

\begin{figure*}[t]
  \centering
  \includegraphics[width=\textwidth]{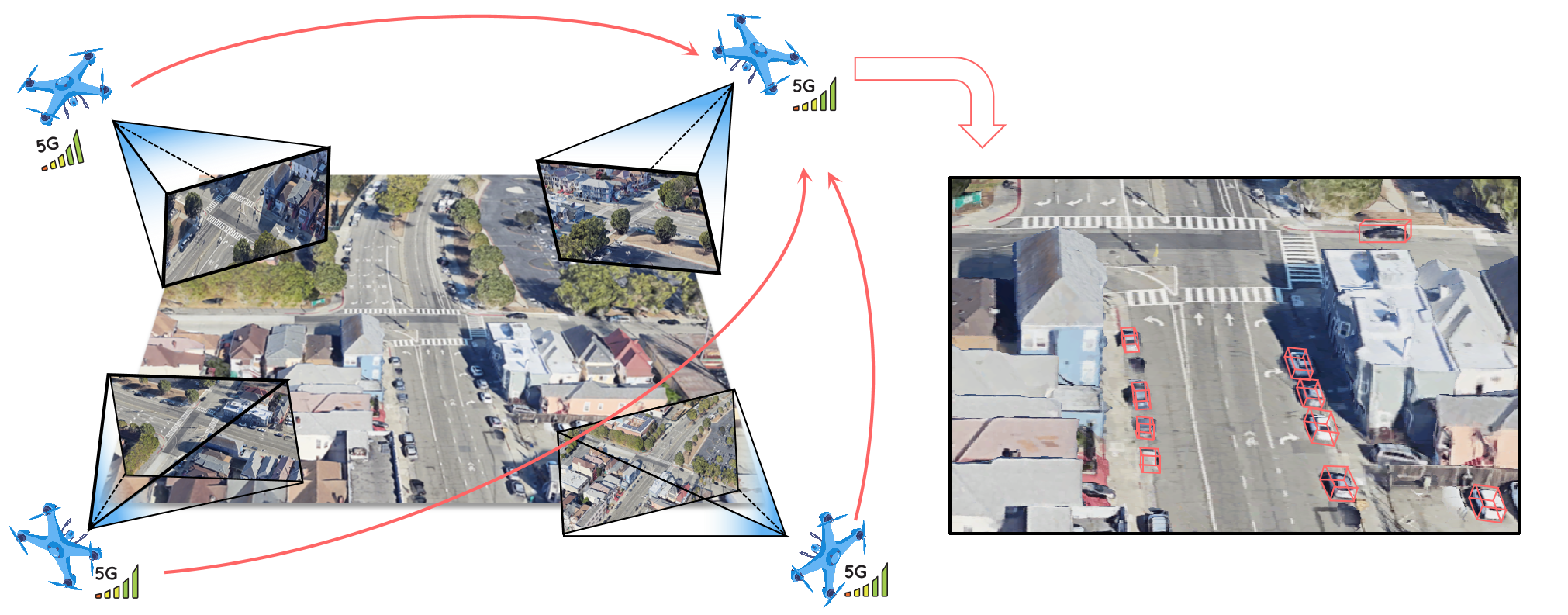}
  \caption{The illustration of multi-UAV collaborative 3D object detection.}
  \label{fig:intro1}
\end{figure*}

Multi-UAV collaborative 3D object detection \cite{zhu2020multi,zong2023temporal,hu2022where2comm} can perceive and understand complex environments by employing cooperative strategies and integrating complementary information as shown in Fig. \ref{fig:intro1}, which has experienced significant advancements in recent years. Compared to single-UAV, multi-UAV collaborative 3D object detection broadens the observation scope and improves the robustness to occlusions, whose applications encompass traffic monitoring \cite{butilua2022urban}, delivery services \cite{li2023drone}, agricultural management \cite{dutta2020application}, and aerial photography \cite{harrigan2024thermography}.

Current multi-UAV collaborative perception methods \cite{xu20213d,men2021cooperative} adopt a multi-stage technological approach. Specifically, it begins with 2D object detection in each view, followed by cross-image object association, and finally encompasses collaborative 3D localization of the identified objects. The multi-stage method could bring about the high complexity of systems and the propensity for error accumulation. In contrast, the Bird's Eye View (BEV) method \cite{philion2020lift} is a novel end-to-end paradigm for collaborative perception in autonomous driving scenarios, where image features from different UAVs are initially extracted, then mapped to a unified 3D coordinate system. After feature fusion, BEV features are obtained which are subsequently used for 3D object detection. This end-to-end approach directly optimizes the entire model, facilitating the learning of global information and enhancing the accuracy of 3D object detection.

Nevertheless, the BEV approach cannot be effectively applied to the UAV remote sensing scenarios due to the difficulty in achieving precise and consistent feature mapping, which is essential for multi-UAV collaborative 3D object detection. Firstly, BEV-based methods map 2D image features to 3D space by predicting pixel-wise categorical mapping probability distribution, which determines the weighted pixel feature at each categorical bin along the perspective ray. In aerial contexts, the high flight altitude of UAVs leads to an extremely broad observation range. The usual response strategy is increasing mapping categories, which actually complicate the attainment of accurate probabilities for each categorical mapping bin, thereby severely impacting the perception tasks in the BEV space.
Secondly, unlike different sensors observing without significant overlapping ranges in original BEV perception, multi-UAV collaborative 3D object detection requires multiple UAVs to observe the same scene, with significant overlap between different perspectives. Substantial perspective variations lead to inadequate generalization capabilities of the feature mapping module for each perspective. During collaboration process, information from distinct UAVs encounters difficulties in constructing a consistent 3D spatial representation. The superimposition of erroneous mapping features onto accurate mapping features distorts BEV representations.

To address the challenges in achieving precise and consistent feature mapping, we utilize the UAV aerial perspective prior to narrow the feature mapping probability distribution and leverage the UAV collaborative observation characteristic to jointly learn feature mapping relationships across multiple views. Specifically, this paper presents a  innovative multi-\textbf{U}AV \textbf{C}ollaborative 3D Object \textbf{D}etection \textbf{Net}work called \textbf{UCDNet}, consisting of a
Ground-Prior-Guided Feature Mapping (GFM) module and a Homologous Point Geometric Consistency Loss (HPL). Firstly, considering objects in the scene are located on the ground, the GFM module first calculates the ground depth for each feature pixel based on ground altitude and camera parameters, then sets a narrow possible range of the feature mapping position near the ground, and finally utilizes a neural network to estimate the probability distribution for the feature mapping classification task. GFM explicitly make use of the ground as a strong prior to provide the reference for more accurate and generalizable feature mapping, which is a simple and effective way. Secondly, motivated by the fact that scenes captured by different UAVs have certain overlapping areas, the HPL begins with extracting homologous points on multi-view images, then calculates their positions in 3D space based on the feature mapping probability distribution, and finally uses the consistency of the 3D spatial positions of the homologous points as an auxiliary self-supervised signal. HPL directly utilizes the relationship between different views and their underlying scene geometry to enforce the learning of a unified and precise 3D space. Independent feature mappings from different UAVs have achieved better results through collaborative learning. Thirdly, we design a dataset rich in scenarios called "AeroCollab3D" simulated by CARLA \cite{dosovitskiy2017carla} to validate the effectiveness of the proposed UCDNet.

The main contributions of this paper are summarized as follows:
\begin{itemize}[leftmargin=*,itemsep=0pt,parsep=0pt,topsep=0pt]
\item {This paper innovatively propose UCDNet, which is a end-to-end method to enhance multi-UAV collaborative 3D object detection by precise and consistent feature mapping.}
\item {We design a Ground-Prior-Guided Feature Mapping module which explicitly make use of ground as a strong prior to provide the reference for more accurate and generalizable feature mapping.}
\item {We present a Homologous Points Geometric Consistency loss as an auxiliary self-supervised signal, which directly influences the feature mapping module, strengthening the global consistency of multi-view feature mapping.}
\item {A simulated dataset, “AeroCollab3D”, designed for multi-UAV collaborative 3D object detection, is introduced to validate the effectiveness of the proposed UCDNet.}
\end{itemize}

\section{Related Works}

\textbf{Multi-UAV Collaborative Perception.}
With the development of communication and remote sensing 
technology \cite{yan2023air,sun2023single}, multi-UAV collaborative perception has already become a burgeoning field. The traditional methods generally adopt a multi-stage framework, consisting of detection, association and localization. After conducting the detection and association of the object across multiple UAVs, Xu et.al \cite{xu20213d} establish system of equations between the object point and pixels on different images according to the perspective projection imaging principle, and calculate the optimal solution by least squares method. Men et.al \cite{men2021cooperative} produces multiple localization results for the same object from different views, from which a optimal solution is selected. 
However, for multi-stage approach, errors generated at each step can impact the final outcome, and the long processing duration is not conducive to real-time UAV perception. In contrast, multi-UAV collaborative perception based on BEV method is an innovative end-to-end collaborative perception framework. Hu et.al \cite{hu2023aerial} and Chen et.al \cite{hu2022where2comm} respectively explore UAV monocular and collaborative 3D object detection through the construction of a virtual dataset. These methods perform poorly because they are not specifically improved for UAV remote sensing scenarios. More methods for UAV perception are worth further explorations. 

\textbf{Multi-View 3D Object Detection.}
Multi-view 3D object detection is a task of predicting 3D bounding boxes in the global system from multiple images , which includes object’s center coordinate, 3D dimensions and yaw angle. Current approaches could be mainly categorized into two branches, LSS-based \cite{philion2020lift,huang2021bevdet,huang2022bevdet4d,li2023bevdepth,zhou2023matrixvt} and query-based \cite{wang2022detr3d,liu2022petr,li2022bevformer,jiang2023far3d,wang2023exploring,liu2023sparsebev} methods.
LSS-based methods utilize a mapping probability distribution to model mapping uncertainty and project multi-view features into the same BEV space.
CaDDN \cite{reading2021categorical} and BEVDet \cite{huang2021bevdet} propose a fully differentiable end-to-end network which treats feature mapping as a classification task rather than a regression task to predict possibilities at each point along the perspective ray in 3D space. BEVDet4D \cite{huang2022bevdet4d} extends the BEVDet by the temporal modeling and achieves good velocity estimation. BEVDepth \cite{li2023bevdepth} utilizes LiDAR point clouds to supervise feature mapping, but point clouds in UAV scene are often too sparse to provide effective information. A drawback of the LSS-based approach is that as the observation range expands, the number of feature mapping categories increases, leading to insufficient classification capability of the model. Query-based methods utilize attention mechanisms to actively query 2D features from 3D space. DETR3D \cite{wang2022detr3d} samples 2D features from the projected 3D reference points and then conducts local cross attention to update the queries. BEVFormer \cite{li2022bevformer} exploits spatial and temporal information through predefined grid-shaped BEV queries. Far3D \cite{jiang2023far3d} constructs 3D queries through the employment of a 2D object detector and a depth network, significantly expanding the range of 3D object detection. When the perceptual range becomes extensive, query-based methods struggle with handling occlusions and overlaps, leading to poor performance. 

\section{Methodology}

\subsection{Problem Formulation}
In this paper, we develop a Multi-UAV collaborative 3D Object Detection Network called \textbf{UCDNet}. Consider N UAVs in the scene, where the N-th UAV is the ego UAV, let $\mathbf{X} _{i}\in \mathbb{R} ^{3\times H\times W} $ denote the image observation captured by the i-th UAV and $\mathbf{X} _{e}$ denote the image from the ego UAV. $\mathbf{Y}$ represents the corresponding ground-truth supervision. The objective of UCDNet is to maximize the camera-based 3D detection performance :
\begin{align}
\underset{\theta}{\mathrm{argmax}} \quad \xi \left( \Phi_{\theta}\left (  \mathbf{X} _{e},\left \{ \mathcal{F}_{i\to e}, \mathcal{P} _{i\to e} \right \}_{i=1}^{N-1}  \right ),\mathbf{Y}\right ),
\end{align}
where $\xi (\cdot, \cdot )$ denotes the perception evaluation metric, $\Phi(\cdot ) $ denotes the multi-UAV collaborative perception network parameterized by $\theta $ and . $\mathcal{F}_{i\to e}\in \mathbb{R} ^{c\times h\times w}$ and $\mathcal{P}_{i\to e}\in \mathbb{R} ^{d\times h\times w}$ is the feature and probability message transmitted from the i-th agent to the ego agent.

\subsection{Overview}
\begin{figure*}[t]
  \centering
  \includegraphics[width=\textwidth]{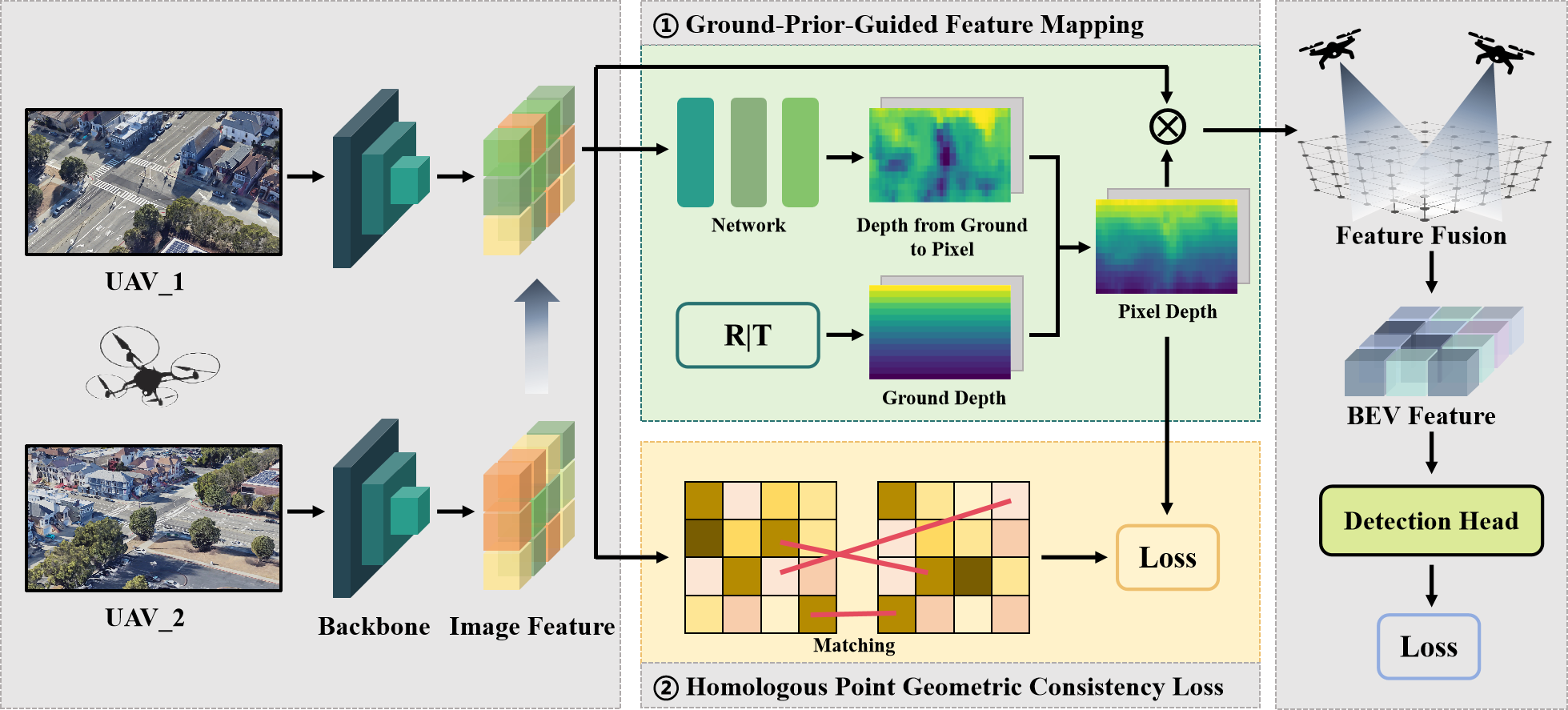}
  \caption{The overall framework of UCDNet, where Ground-Prior-Guided Feature Mapping explicitly utilize ground as a strong prior to provide the reference for more accurate and generalizable feature mapping and Homologous Point Geometric Consistency Loss is proposed as a auxiliary self-supervision, which directly influences the feature mapping network, strengthening the global consistency of multi-view perception.}
  \label{fig:overview}
\end{figure*}

 The overall framework of our proposed multi-UAV collaborative 3D object detection model is illustrated in Fig. \ref{fig:overview}, consisting of four steps: image encoder, feature mapping, feature fusion and 3D detection head. In image encoder, the framework initially extracts the image feature $\mathcal{F}_i$ and its mapping probability distribution $\mathcal{P}_i$ through the backbone for each UAV, whose weights are shared across multiple agents, followed by the subordinate UAVs transmitting messages to the ego UAV. In feature mapping, the mapping probability distribution is utilized to guide the pixel features from N UAVs backproject into each spatial position along the perspective ray in a unified world coordinate system. Then, feature fusion subsequently acquires the fused BEV feature by a collapse and FPN module. Finally, a task-specific head is built upon the BEV feature and perform 3D object detection. 

\subsection{Ground-Prior-Guided Feature Mapping}

\begin{wrapfigure}{R}{0.4\textwidth} 
	\centering
	\includegraphics[width=0.4\textwidth]{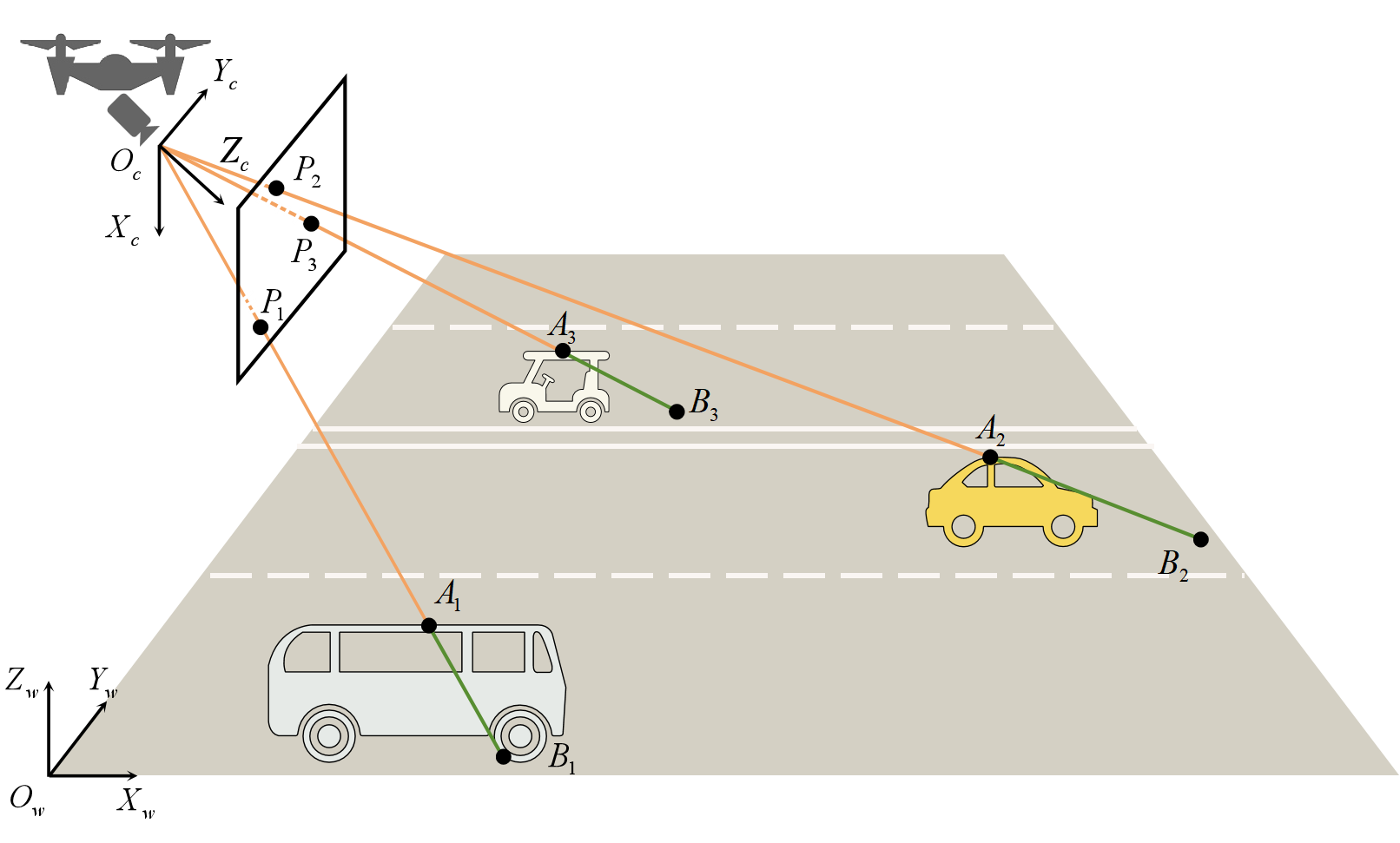}
	\caption{UAV perspective diagram.} 
	\label{fig:depthone}
    \vspace{-10mm}
\end{wrapfigure}

\begin{figure}[t]
    \centering
    \begin{minipage}[b]{0.62\textwidth}
        \centering
        \includegraphics[width=\textwidth]{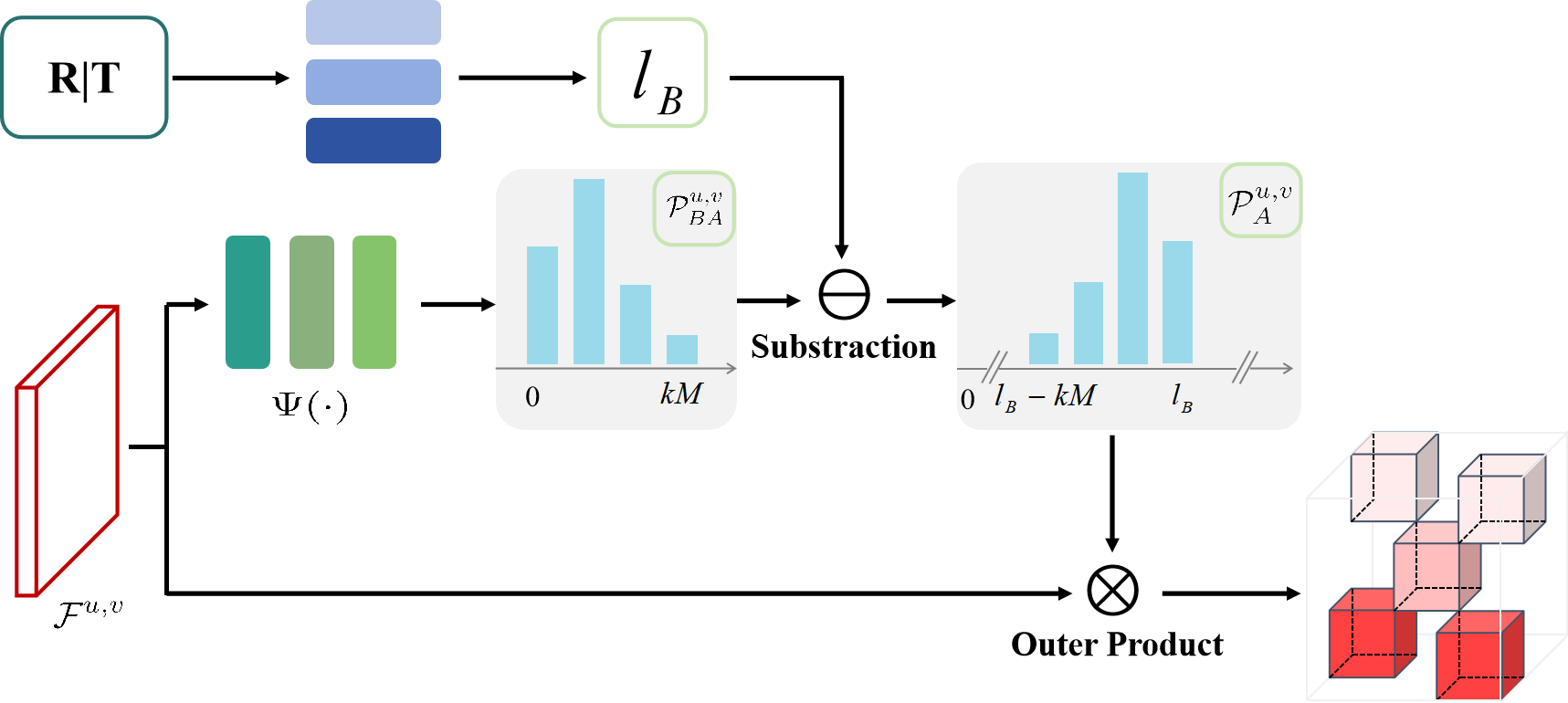}
        \caption{Schematic of the GFM module’s operational flow.}
        \label{fig:groundmodule}
    \end{minipage}
    \hfill
    \begin{minipage}[b]{0.34\textwidth}
        \centering
        \includegraphics[width=\textwidth]{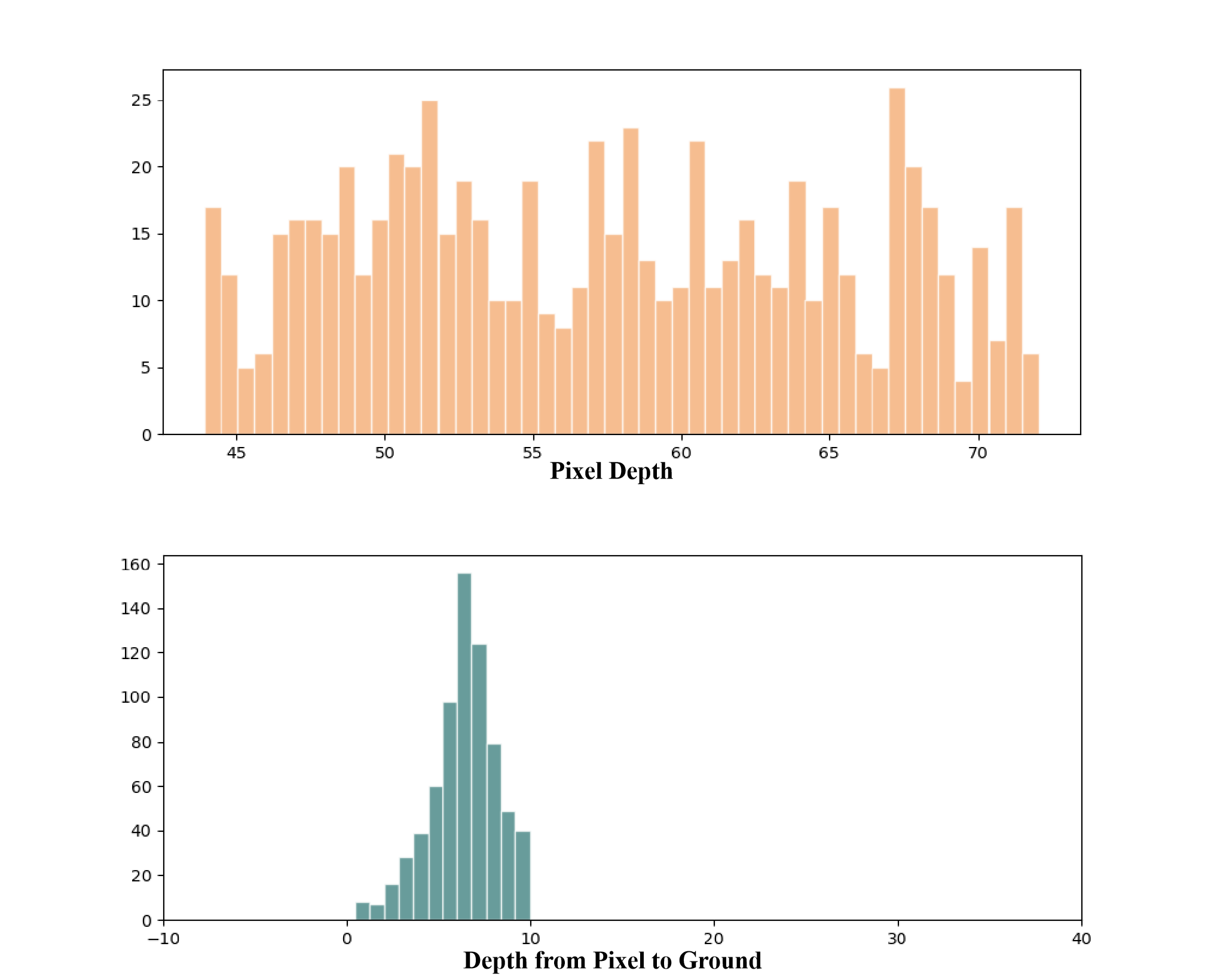}
        \caption{Histogram of pixel depth and depth from ground to pixel.}
        \label{fig:depthtwo}
    \end{minipage}
\end{figure}

The precise mapping of pixel features from the image domain to the spatial domain is crucial for achieving multi-UAV collaborative 3D object detection. BEV methods utilize the probability distribution of pixel depth $\mathcal{P}_{A}^{u,v}$ to guide the weighted pixel feature $\mathcal{F}^{u,v}$ in mapping at each point along the perspective ray. The expansive observation range of UAVs makes it difficult to map pixel features to their correct 3D positions with the maximum probability. To improve the accuracy of feature mapping, we propose GFM module, which introduces the ground depth as a strong prior to provide the reference for feature mapping, with its methodology illustrated in Fig. \ref{fig:groundmodule}


{\bf Definitions.}
As illustrated in Fig. \ref{fig:depthone}, $O_{c}-X_{c}Y_{c}Z_{c}$ and $O_{w}-X_{w}Y_{w}Z_{w}$ are the camera coordinate system and world coordinate system. Within the camera imaging model, a ray emanates from the camera center $O_{c}$ towards a pixel $P=(u, v)$, intersecting with an object in 3D space at point $A$, and further extending to intersect with the ground at point $B$. In our definition, the length of $O_{c}A$ along $Z_{c}$ is the pixel depth $l_{A}$, collectively forming the image depth $\mathcal{D}$. The length of $O_{c}B$ along $Z_{c}$ is the ground depth prior $l_{B}$, and the length of $BA$ along $Z_{c}$ is the depth from the ground to the pixel $l_{AB}$.

{\bf Ground Depth Calculation.}
The intrinsic parameters of the camera are denoted by $\mathbf{K}\in \mathbb{R}^{3\times 3}$, while $\mathbf{R}\in \mathbb{R}^{3\times 3}$ and $\mathbf{T}\in \mathbb{R}^{3\times 1}$ respectively represent the rotation and translation matrices of the extrinsic parameters, which are known from the dataset. Then the ground depth prior $l_B$ can be calculated as follows and detailed derivation can be found in the supplementary material.
\begin{align}
l_{B}=\frac{z_{w}-[\mathbf{R}^{-1}(-\mathbf{T})]_{3}}{[\mathbf{R}^{-1}\mathbf{K^{-1}}]_{31}u+[\mathbf{R}^{-1}\mathbf{K^{-1}}]_{32}v+[\mathbf{R}^{-1}\mathbf{K^{-1}}]_{33}} .
\end{align}

{\bf Depth Estimation Guided by Ground Depth Prior.}
UAVs possess a unique aerial perspective, which results in a distinctly narrower distribution range for $l_{BA}$ compared to $l_{A}$, as illustrated in Fig. \ref{fig:depthtwo}. Based on the ground depth prior $l_{B}$, our method firstly estimates the probability distribution of $l_{BA}$ represented by $ \mathcal{P}_{BA}^{u,v}$ using a simple network $\Psi(\cdot )$.
\begin{align}
\mathcal{P}_{BA}^{u,v} = \Psi \left ( \mathcal{F}^{u,v}   \right ) = \left \{ \mathrm{P}\left ( kd \right )  \right \} _{k = 1}^{M} ,
\end{align}
where $l_{BA}$ is divided into M depth bins, each with a length of d. Then the probability distribution of $l_{A}$ represented by $ \mathcal{P}_{A}^{u,v}$ can be calculated by the conversion formula as follows:
\begin{align}
\mathcal{P}_{A}^{u,v} = \left \{ \mathrm{P}\left ( l_B-kd \right )  \right \} _{k = 1}^{M}. 
\end{align}

{\bf Feature Mapping Process.}
Finally, the pixel feature $\mathcal{F}^{u,v}$ is mapped into the camera coordinate system according to the probability distribution $\mathcal{P}_{A}^{u,v}$ by outer product $\otimes$, and transformed into a unified world coordinate system represented by $\mathfrak{T} _{c}^{w}$.
\begin{align}
\mathfrak{T} _{c}^{w} \left ( \mathcal{F}^{u,v} \otimes \mathcal{P}_{A}^{u,v}  \right ) .
\end{align}
Ground-prior-guided feature mapping is an economical and effective method to enhance the feature mapping accuracy by reducing the mapping range. We directly and explicitly take advantage of ground depth as a precise reference and strong prior to improve the robustness of feature mapping from the UAV perspective.  The mapping positions become more reasonable when it is constrained near the real ground.

\subsection{Homologous Point Geometric Consistency Loss}

In multi-UAV collaborative 3D object detection task, there are numerous overlapping observation areas among different UAVs, the image features $\mathcal{F}_i$ from which are mapped into a unified 3D space. If the feature mapping from a certain perspective is inaccurate, it will affect the feature fusion and reduce the accuracy of 3D object detection. Therefore, enhancing the consistency of multi-view feature mapping is particularly crucial. In this paper, we design an innovative homologous point geometric consistency loss, where independently feature mappings from different UAVs supervise each other, collaboratively learning a consistent mapping relationship.

\begin{figure}[t]
    \centering
    \begin{minipage}[b]{0.62\textwidth}
        \centering
        \includegraphics[width=\textwidth]{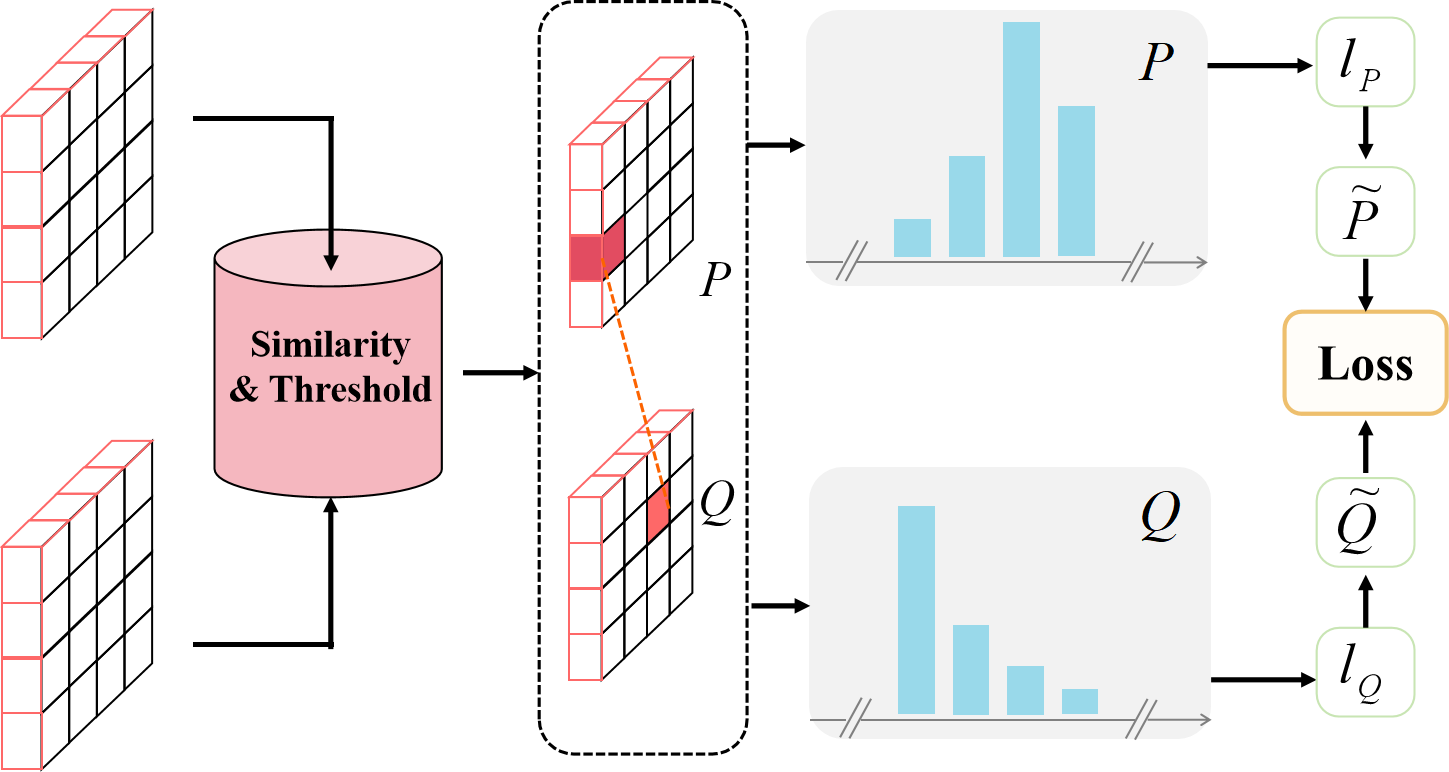}
        \caption{Schematic of the HPL’s operational flow.}
        \label{fig:points2}
    \end{minipage}
    \hfill
    \begin{minipage}[b]{0.34\textwidth}
        \centering
        \includegraphics[width=\textwidth]{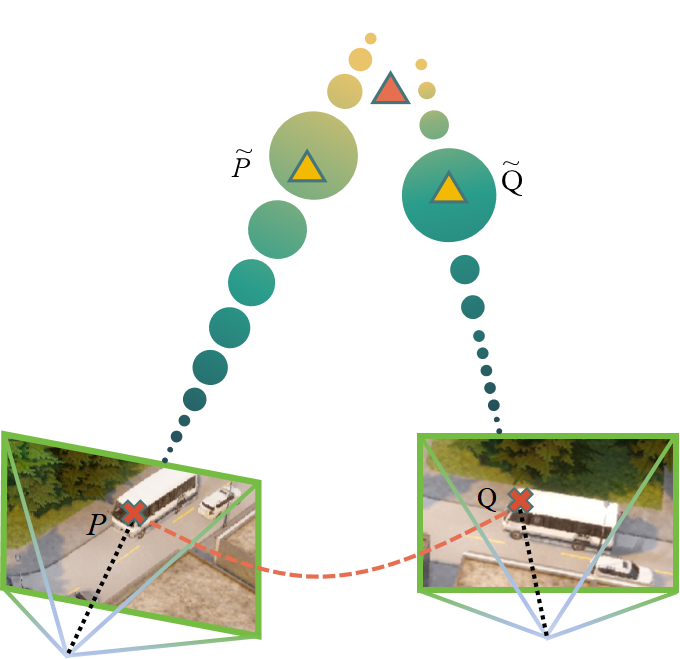}
        \caption{Homologous point geometric consistency.}
        \label{fig:points}
    \end{minipage}
\end{figure}

Homologous points \cite{hartley2003multiple} denote the pixels in different images, which represent the same physical point, as illustrated in Fig. \ref{fig:points}. Homologous points posses geometric consistency because they can backproject to the same 3D position in the world coordinate system. Given the overlapping areas captured by each UAV, we can extract homologous points from multi-view image features, utilizing the geometric consistency loss of these points as an auxiliary supervisory signal to get more consistent feature mapping. Considering the background area lacks supervisory signals, which leads to the inherently impossibility of accurate feature mapping, we only extract homologous points from object regions to fully leverage the potential of homologous point supervision. The detailed steps HPL are as follows and inllustrated in Fig. \ref{fig:points2}.

Assuming that given a feature image pair $\left ( \mathcal{F}_i, \mathcal{F}_j \right )$, we aims to obtain pairs of Homologous pixels $P\in \mathcal{F}_i$ and $Q\in \mathcal{F}_i$. Firstly, We use a simple network to obtain a predicted object area mask. This predicted mask is then used to constrain the extraction range of the homologous points. Secondly, we compute cross-attention between two feature images to acquire the correlation between any two points. Thirdly, we set both a numerical threshold and a correlation threshold, and only those homologous points that satisfy both thresholds are retained. The formula of our proposed loss function are as follows:
 
\begin{align}
L_{\text{consistency}} = \frac{1}{N} \sum_{k=1}^{N} \lVert \tilde{P}^{(k)} -\tilde{Q}^{(k)} \rVert_2^2.
\end{align}

We use the Euclidean distance between the 3D positions $\tilde{P}$, $ \tilde{Q}$ of homologous points $P$, $Q$ as the loss function. The transformation between homologous points and their corresponding 3D positions is as follows:
\vspace{-5pt} 
\begin{align}
\tilde{P} &= \mathbf{R}_{P}^{-1}(\mathbf{K}_{P}^{-1}\bar{P}z_{P}-\mathbf{T}_{P}) \quad \tilde{Q} = \mathbf{R}_{Q}^{-1}(\mathbf{K}_{Q}^{-1}\bar{Q}z_{Q}-\mathbf{T}_{Q}),
\end{align}
\vspace{-7pt} 
where $\bar{P}$, $\bar{Q}$ denote the homogeneous form of $P$, $Q$ and $z_{P}$, $z_{Q}$ are the weighted pixel depth of $P,Q$.

Our homologous points geometric consistency loss explicitly exploits the relationship between different images and their underlying scene geometry, enforcing the learning of a global and geometrically precise solution, thereby enhancing the model's understanding of the 3D scene and improving object detection effect without exploiting extra computational cost during inference. Moreover, our homologous points geometric consistency loss directly affects the feature mapping network and image features rather than downstream BEV features, enhancing the accuracy of 3D object detection while accelerating model convergence.

\subsection{AeroCollab3D Dataset}

\begin{figure}[htbp]
    \centering
    \begin{minipage}[t]{0.33\textwidth}
        \centering
        \includegraphics[width=\textwidth]{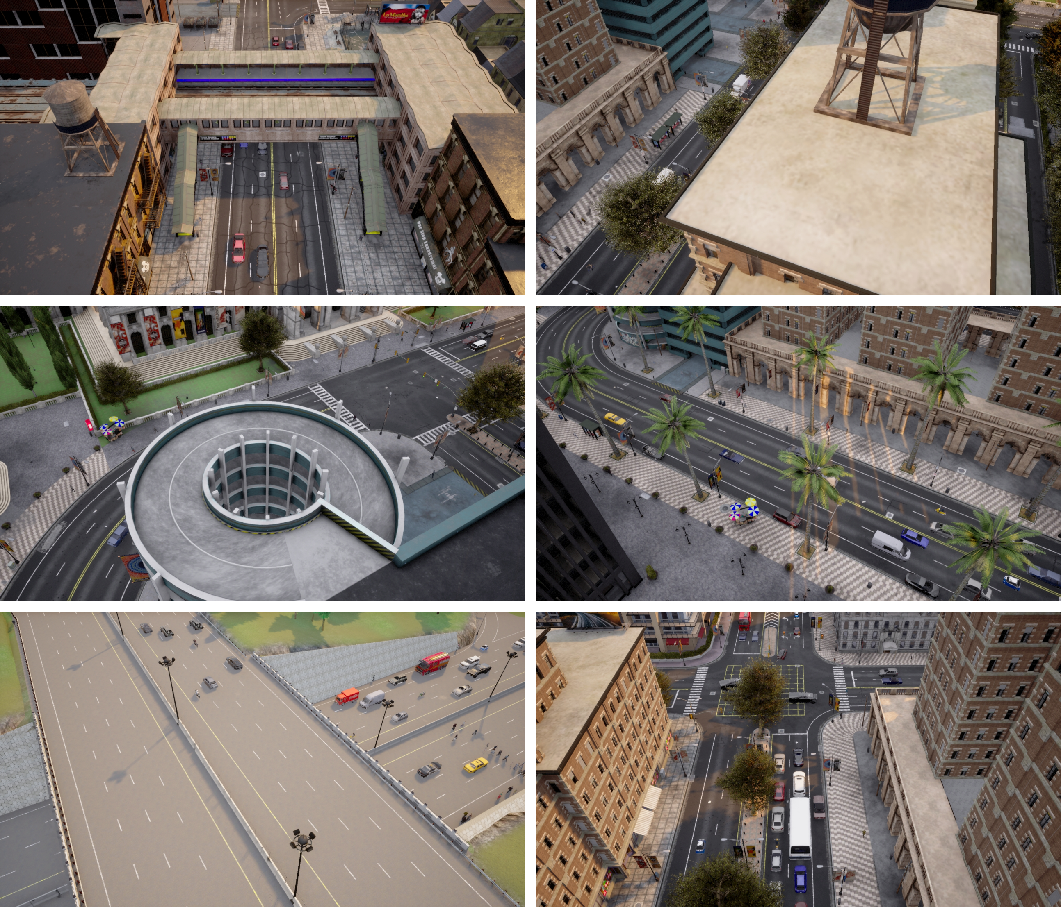}
        \caption{Challenging scenes in AeroCollab3D dataset.}
        \label{fig:datasetshow}
    \end{minipage}
    \hfill
    \begin{minipage}[t]{0.65\textwidth}
        \centering
        \vspace{-115pt} 
        \captionof{table}{Comparison of CoPerception-UAVs and AeroCollab3D.}
        \begin{adjustbox}{width=\textwidth}
        \begin{tabular}{@{}lcc@{}}
            \toprule
            \textbf{Attribute} & \textbf{CoPerception-UAVs} & \textbf{AeroCollab3D} \\
            \midrule
            Number of Maps       & 3                & \textbf{8} \\
            Number of Agents     & 5                & \textbf{6} \\
            Number of Sensors    & 5                & 1 \\
            Flight Height        & 60m              & 50m \\
            Object Categories & 1                & \textbf{4} \\
            Capture Frequency    & 0.25Hz           & \textbf{2Hz} \\
            Number of Samples    & \textbf{5276}    & 3200 \\
            Image Resolution     & 800 $\times$ 450 & \textbf{1600 $\times$ 900} \\
            \bottomrule
        \end{tabular}
        \end{adjustbox}
        \label{tab:dataset_comparison}
    \end{minipage}
\end{figure}

To better implement multi-UAV collaborative 3D object detection, we design a large-scale UAV-based collaborative perception dataset called {\bf AeroCollab3D}, simulated by CARLA \cite{dosovitskiy2017carla}, including 8 maps, 3.2K samples, 19.2K UAV images and 218K 3D boxes across 4 categories. We set 6 collaborative UAVs flying at the height of 50m, each of which equips with only 1 camera at a pitch degree of -45$^\circ$ to simulate the real situation. The image resolution is 1600 $\times$ 900, and the detection range is 110m $\times$ 110m. The sample data is collected randomly at 2 Hz. The dataset employs the same evaluation metrics as the nuScenes dataset. Tab. \ref{tab:dataset_comparison} compares AeroCollab3D with the existing multi-UAV 3D object detection dataset CoPerception-UAVs. The proposed AeroCollab3D has more object categories, larger image resolutions, higher capture frequency which is beneficial for temporal perception, and richer scenarios with high-rise buildings and roadside trees leading to observation occlusions, shown in Fig. \ref{fig:datasetshow}. More details about AeroCollab3D can be found in the supplemental material.

\section{Experiments}


\subsection{Datasets and Evaluation Metrics}
We evaluate our approach on CoPerception-UAVs \cite{hu2022where2comm} and AeroCollab3D dataset. CoPerception-UAVs is a multi-UAV collaborative perception dataset co-simulated by AirSim \cite{shah2018airsim} and Carla, including 3 map, 1 category of object and 5 UAVs, each equipped with 5 cameras. The input image size is 800 $\times$ 450 and the perception range is 200m $\times$ 350m. AeroCollab3D is our proposed dataset mentioned in Sec.3.  

The 3D object objection evaluation metrics are mean Average Precision (mAP), mean Absolute Trajectory Error (mATE), mean Absolute Scale Error (mASE), and mean Absolute Orientation Error (mAOE), where 0.5, 1.0, 2.0, 4.0 are used as the threshold of object center distance to judge true positives. To quantitatively evaluate the impact of UCDNet on feature mapping, we design an evaluation metric mean Feature Mapping Score (mFMS). Centering on the object GT position, a cubic region of $4 \times 4 \times 4$ is defined. The probability values of all features projected within this region are summed for FMS and each object scores sre averaged to obtain the mFMS. A higher result indicates a greater probability that the object features are correctly mapped to the intended position. 

\subsection{Implementation Details}
In our experimental setup, we select BEVDet4D \cite{huang2022bevdet4d} as our baseline and CenterPoint \cite{yin2021center} is served as the 3D object detection head. Our method is trained by AdamW optimizer \cite{loshchilov2017decoupled} with a learning rate of 2 $\times 10^{-4}$ . The models undergo 40 epochs of training with a batch size of 8. Only key frames are used during both training and inference. All experiments are conducted on an RTX-3090 GPU with PyTorch version 1.8.3.

\subsection{Main Results}


\textbf{Results on AeroCollab3D dateset.} We compare the proposed UCDNet with previous state-of-the-art vision-based 3D detectors on the AeroCollab3D test set. As shown in Tab. \ref{tab:methods_comparison}, UCDNet shows superior performance on mAP, mATE, mASE, and mAOE metrics when adopting ResNet50 backbone with nuImages pretraining. Compared with the baseline BEVDet4D, UCDNet has considerable improvements of 4.7$\%$ mAP. The mATE of UCDNet is 4.2$\%$ better than BEVDet4D, indicating that our method is able to improve the accuracy of localization. Compared to mainstream methods that enhance detection results through the use of multiple frames, such as HOP \cite{zong2023temporal} and StreamPETR \cite{wang2023exploring}, UCDNet employs the baseline setting of utilizing only two frames but surpasses the former in detection outcomes. In contrast to the BEVdepth method, which employs point cloud data as a supervisory signal, our approach, utilizing solely optical images, achieves superior results. This demonstrates that the method proposed in this study effectively enhances the accuracy of depth estimation to improve detection performance.

\textbf{Results on CoPerception-UAVs dataset.}
In our implementation, we exclusively deploy three UAVs for collaboration perception, as our observations indicate that the remaining two UAVs do not share any overlapping fields of view with the ego agent, thus offering no enhancement to the perception outcomes of the main drone. We compare our UCDNet with other state-of-the-art methods like Where2comm, BEVDet on CoPerception-UAVs test set. As shown in Tab. \ref{tab:methods_comparison}, our proposed method is improved by 10$\%$ mAP on CoPerception-UAVs compared with baseline, with an obvious effect. The UAV flight altitude of CoPerception-UAVs is 60m, which is greater than the AeroCollab3D data set, so our method is more obvious than baseline, which proves the significant advantage of our method in UAV 3D object detection. To ensure a fair comparison with the where2comm methodology, we merely integrated its collaboration strategy into our baseline, while the backbone network and detection head remained unchanged. Experimental outcomes demonstrate that UCDNet significantly outperforms where2comm in terms of detection results.


\begin{table}[!h]
    \centering
    \caption{Comparing with the state-of-the-art on AeroCollab3D and CoPerception-UAVs test set. $^\star$ denotes our baseline model, whereas $^\dagger$ indicates that we integrates the collaboration module proposed by Where2comm into the baseline as the comparative method. $-$ represents CoPerception-UAVs dataset cannot be applied to StreamPETR and HoP due to the lack of point cloud information.}
    \label{tab:methods_comparison}
    \resizebox{\textwidth}{!}{
    \begin{tabular}{@{}lcccccccccc@{}}
        \toprule
        \multirow{2}{*}{\textbf{Method}} & \multicolumn{4}{c}{\textbf{AeroCollab3D}} & & \multicolumn{4}{c}{\textbf{CoPerception-UAVs}} \\
        \cmidrule(lr){2-5} \cmidrule(lr){7-10}
        & \textbf{mAP$\uparrow$} & \textbf{mATE$\downarrow$} & \textbf{mASE$\downarrow$} & \textbf{mAOE$\downarrow$} & & \textbf{mAP$\uparrow$} & \textbf{mATE$\downarrow$} & \textbf{mASE$\downarrow$} & \textbf{mAOE$\downarrow$} \\
        \midrule
        BEVDet \cite{huang2021bevdet}                  & 0.584 & 0.478 & 0.141 & 0.453 & & 0.334 & 1.010 & 0.169 & 1.574 \\
        BEVDet4D$^\star$ \cite{huang2022bevdet4d}              & 0.592 & 0.459 & 0.119 & \textbf{0.363} & & 0.347 & 0.992 & 0.170 & 1.099 \\
        BEVLongTerm \cite{huang2022bevdet4d}           & 0.356 & 0.585 & 0.244 & 0.583 & & 0.255 & 1.023 & 0.201 & 1.823 \\
        Where2comm$^{\dagger}$ \cite{hu2022where2comm} & 0.402 & 0.498 & 0.201 & 0.462 & & 0.298 & 0.997 & 0.178 & 1.084 \\
        StreamPETR \cite{wang2023exploring}            & 0.456 & 0.631 & 0.122 & 0.458 & & 0.312 & 1.002 & 0.185 & 1.694 \\
        HoP \cite{zong2023temporal}                    & 0.180 & 0.812 & 0.248 & 1.066 & & $-$   & $-$   & $-$   & $-$\\
        BEVDepth \cite{li2023bevdepth}                 & 0.601 & 0.438 & 0.153 & 0.451 & & $-$   & $-$   & $-$   & $-$ \\
        \textbf{UCDNet}                                & \textbf{0.639} & \textbf{0.417} & \textbf{0.112} & 0.388 & & \textbf{0.434} & \textbf{0.853} & \textbf{0.164} & \textbf{1.039} \\
        \bottomrule
    \end{tabular}
    }
\end{table}

\begin{figure}[!h]
  \centering
  \includegraphics[width=\linewidth]{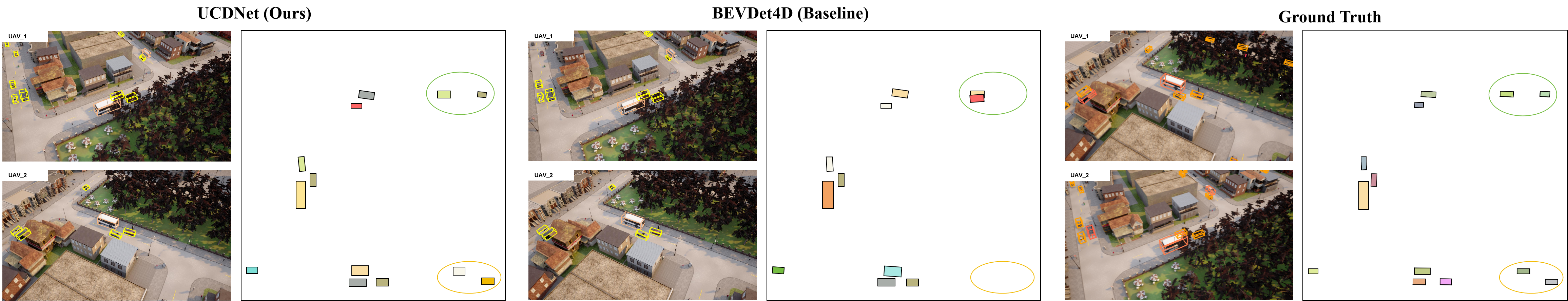}
  \caption{Visualization Results of AeroCollab3D dataset.}
  \label{fig:visualpicture}
\end{figure}

\textbf{Visualization Results.}
Fig. \ref{fig:visualpicture} shows the visualization results of our method on AeroCollab3D 
 dataset. UCDNet's detection bounding boxes offer better fitting to the objects and are capable of collaboratively detecting small and occluded objects, demonstrating both accuracy and robustness.

\vspace{-10pt} 

\subsection{Ablation Study $\&$ Analysis}
\textbf{Modules Analysis within the multi-UAV collaborative Perception Framework:}
As described in the Sec.3, the proposed framework contains two key components: GFM and HPL. We conduct a series of ablation experiments on the AeroCollab3D dataset to analyze the contribution of each module and the experimental results are depicted in Tab. \ref{tab:Ablation}. After utilizing GFM module, our method significantly enhances 3.6$\%$ mAP and decreases 5.8$\%$ mATE compared to baseline, which demonstrates the reliability and suitability of GFM module in multi-UAV collaborative 3D object detection. Additionally, adding HPL to the baseline during training period gain the increase of 2.2$\%$ mAP, which is a effective and economical way to enhance the accuracy in multi-UAV collaborative 3D object detection. Furthermore, it is noteworthy that our method performs better on stricter evaluation metrics. For instance, after adding both the GFM and HPL modules, our method gains a 2.7$\%$ improvement in AP$_{4.0}$ but a 10.0$\%$ improvement in AP$_{0.5}$. 

\begin{table}[h!]
\centering
\caption{Ablation experiments results.}
\label{tab:Ablation}
\resizebox{\textwidth}{!}{
\begin{tabular}{
    cc
    S[table-format=1.4]
    S[table-format=1.4]
    S[table-format=1.4]
    S[table-format=1.4]
    S[table-format=1.4]
    S[table-format=1.4]
    S[table-format=1.4]
    S[table-format=1.4]
    S[table-format=1.4]
}
\toprule
\multicolumn{2}{c}{\textbf{Module}} & \multicolumn{8}{c}{\textbf{Metrics}} \\
\cmidrule(lr){1-2} \cmidrule(lr){3-11}
\textbf{GFM} & \textbf{HPL} & {\textbf{mAP}} $\uparrow$ & {$\textbf{AP}_{0.5}$} $\uparrow$ & {$\textbf{AP}_{1.0}$} $\uparrow$ & {$\textbf{AP}_{2.0}$} $\uparrow$ & {$\textbf{AP}_{4.0}$} $\uparrow$ & {\textbf{mATE}} $\downarrow$ & {\textbf{mASE}} $\downarrow$ & {\textbf{mAOE}} $\downarrow$ & {\textbf{mFMS}}$\uparrow$\\
\midrule
\ding{55} & \ding{55} & 0.592 & 0.264 & 0.525 & 0.732 & 0.767 & 0.459 & 0.119 & 0.363  & 3.916\\
\checkmark & \ding{55} & 0.628 & 0.359 & 0.627 & 0.749 & 0.779 & 0.401 & 0.125 & 0.407  & 5.763\\
\ding{55} & \checkmark & 0.614 & 0.331 & 0.615 & 0.738 & 0.773 & 0.411 & 0.115 & 0.404  & 4.623\\
\checkmark & \checkmark & 0.639 & 0.353 & 0.642 & 0.766 & 0.794 & 0.417 & 0.112 & 0.388  & 6.189\\
\bottomrule
\end{tabular}
}
\end{table}

\begin{figure}[!h]
  \centering
  \includegraphics[width=\linewidth]{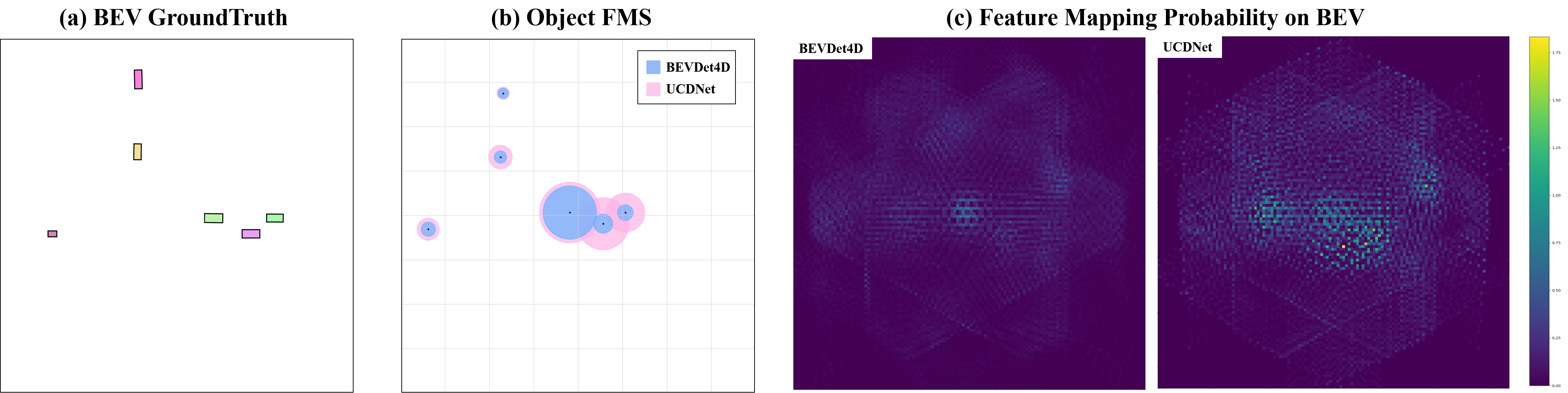}
  \caption{Visualization results about feature mapping.}
  \label{fig:mapping}
\end{figure}

\textbf{Discussion on Feature Mapping.}
As depicted in Tab. \ref{tab:Ablation},UCDNet improves 2.273 mFMS compared to the baseline. Fig. \ref{fig:mapping}(b) uses object FMS as the radius to draw a circle, which can intuitively reflect the improvement of UCDNet on feature mapping. Fig. \ref{fig:mapping}(c) uses colors to represent the sum of probabilities obtained at each location on the BEV feature, where object area is more obvious. Through the discussion above, the proposed UCDNet can achieve more reliable feature mapping to enhance multi-UAV collaborative 3D object detection.

\begin{figure}[!htb]  
    \begin{minipage}[b]{.32\textwidth} 
        \centering  
        \includegraphics[width=\linewidth]{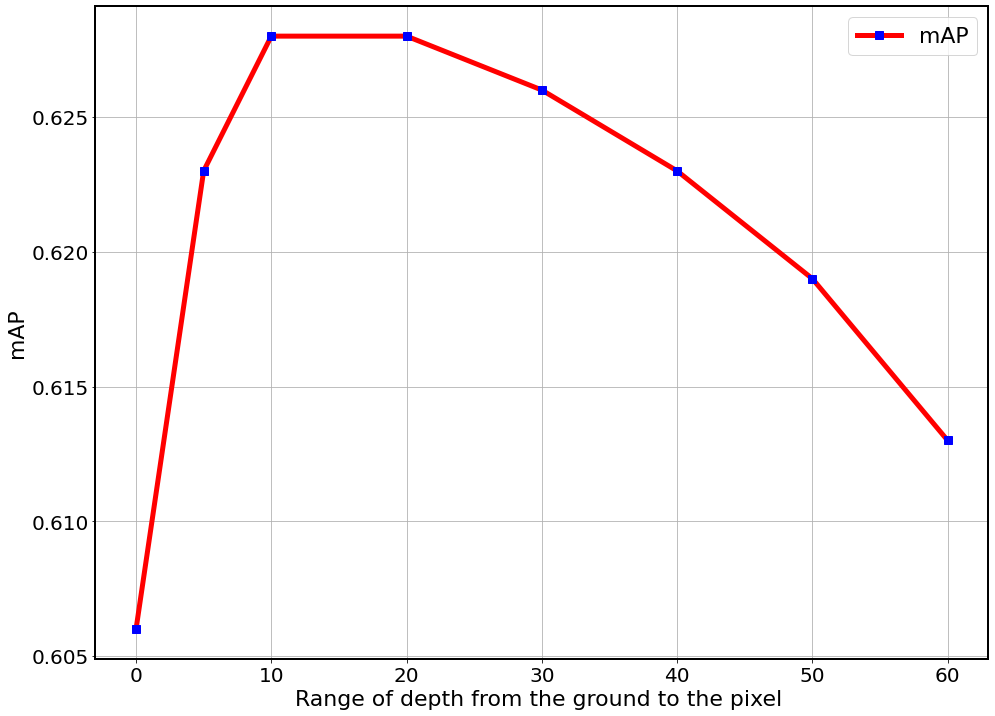} 
        \caption{Detection results of different depth from the ground to the pixe range.}  
        \label{fig:Gplot}  
    \end{minipage}   
    \hfill 
    \begin{minipage}[b]{.32\textwidth} 
        \centering  
        \includegraphics[width=\linewidth]{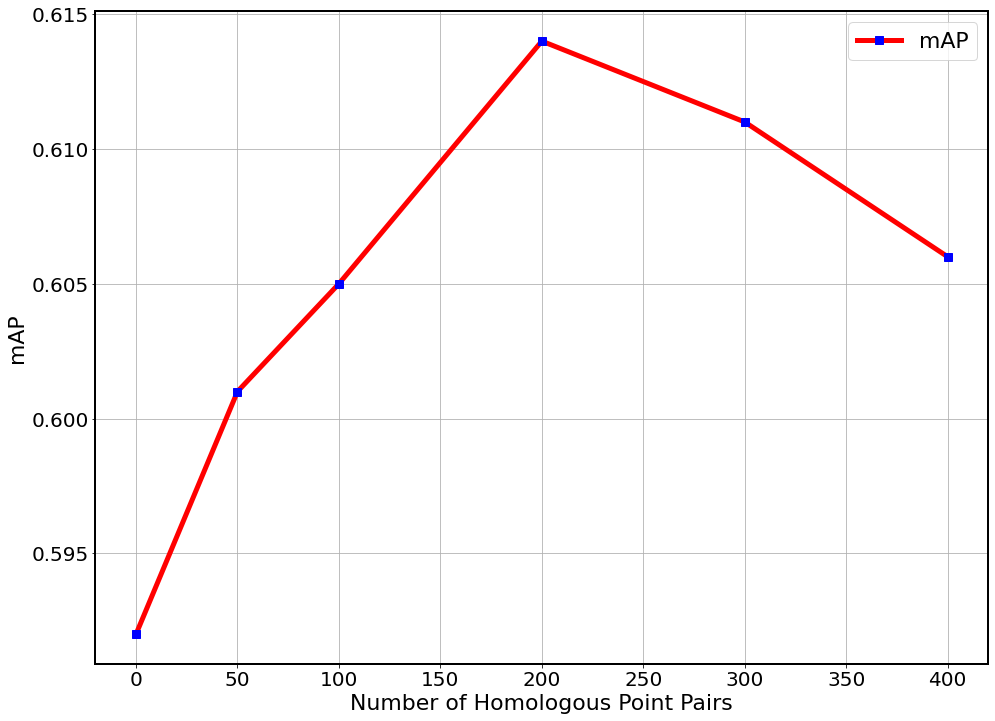} 
        \caption{Detection results of different homologous point pair number.}  
        \label{fig:Hplot}  
    \end{minipage}   
    \hfill 
    \begin{minipage}[b]{.32\textwidth} 
        \centering  
        \includegraphics[width=\linewidth]{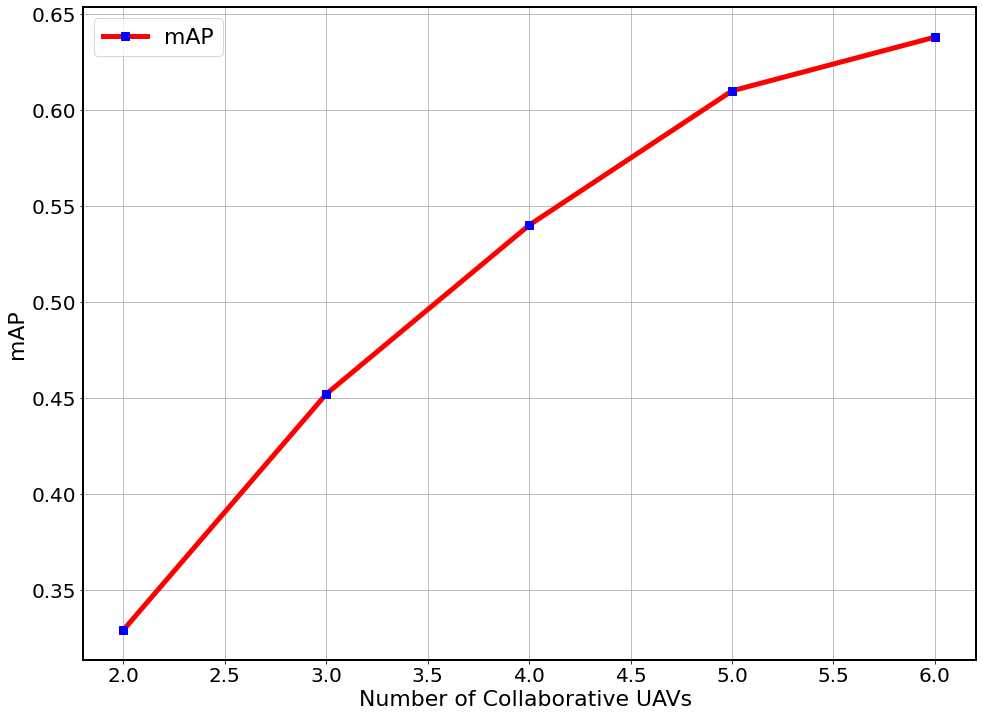} 
        \caption{Detection results of different collaborative UAV numbers.}  
        \label{fig:Uplot}  
    \end{minipage}   
\end{figure}

\textbf{Depth from the Ground to the Pixel Range in GFM Module.}
In GFM module, we set $l_{BA}$ range for categorical depth estimation task. Fig. \ref{fig:Gplot} depicts the variation of detection performance in response to different $l_{BA}$ range settings. $l_{BA}=0$ means we directly use $l_{B}$ as $l_{A}$ without depth estimation, which gains moderate improvement compared to baseline. But if we want to acquire more accurate detection, we should consider object’s scale and camera view to design a suitable $l_{BA}$ range. A too large $l_{BA}$ range will increase the difficulty of the classification task, leading to a decline in detection results.

\textbf{Homologous Point Pair Number in HPL Module.}
The number of homologous point pairs involved in HPL determines the supervisory effectiveness. Smaller numbers cannot provide enough supervision to make multi-view learn consistent mapping relationships, while larger numbers may introduce incorrect Homologous Points which misleads the model's convergence. After investigating the effects of various homologous point pair numbers, as shown in Fig. \ref{fig:Hplot}, we select 200 homologous point pairs per batch in HPL for optimal performance.

\textbf{Collaborative UAV Number Analysis.}
In this experiment, we study the impact of the number of different collaborative UAV on the perceptual outcomes. As Fig. \ref{fig:Uplot} shows, the accuracy of collaborative perception improves with the increase of collaborative UAV number, which proves the rationality multi-UAV collaborative perception.

\vspace{-10pt} 

\section{Conclusion}

In this paper, we propose an unparalleled muti-UAV collaborative 3D object detection network called UCDNet. To address the feature mapping challenge from multi-UAV collaborative perception, the ground depth is explicitly utilized as a strong prior to provide the reference for more accurate and generalizable feature mapping. Besides, we introduce homologous points geometric consistency loss as a auxiliary self-supervision, which strengthens the global consistency of multi-view perception. Additionally, we construct a simulated dataset for multi-UAV collaborative perception task. Experimental results sufficiently demonstrate the superiority of the proposed UCDNet, and ablation studies confirm the effectiveness of the two proposed modules.

\textbf{Limitation and Future Work.} The current research only utilizes simulated datasets to validate the algorithm's effectiveness. In the future, we plan to extend this research to real-world scenarios, investigating issues such as sensor noise, communication delays, etc.
 
\bibliographystyle{unsrt}
\bibliography{neurips_2024}

\begin{thebibliography}{10}

\bibitem{zhu2020multi}
Pengfei Zhu, Jiayu Zheng, Dawei Du, Longyin Wen, Yiming Sun, and Qinghua Hu.
\newblock Multi-drone-based single object tracking with agent sharing network.
\newblock {\em IEEE Transactions on Circuits and Systems for Video Technology}, 31(10):4058--4070, 2020.

\bibitem{zong2023temporal}
Zhuofan Zong, Dongzhi Jiang, Guanglu Song, Zeyue Xue, Jingyong Su, Hongsheng Li, and Yu~Liu.
\newblock Temporal enhanced training of multi-view 3d object detector via historical object prediction.
\newblock In {\em Proceedings of the IEEE/CVF International Conference on Computer Vision}, pages 3781--3790, 2023.

\bibitem{hu2022where2comm}
Yue Hu, Shaoheng Fang, Zixing Lei, Yiqi Zhong, and Siheng Chen.
\newblock Where2comm: Communication-efficient collaborative perception via spatial confidence maps.
\newblock {\em Advances in neural information processing systems}, 35:4874--4886, 2022.

\bibitem{butilua2022urban}
Eugen~Valentin Butil{\u{a}} and R{\u{a}}zvan~Gabriel Boboc.
\newblock Urban traffic monitoring and analysis using unmanned aerial vehicles (uavs): A systematic literature review.
\newblock {\em Remote Sensing}, 14(3):620, 2022.

\bibitem{li2023drone}
Xueping Li, Jose Tupayachi, Aliza Sharmin, and Madelaine Martinez~Ferguson.
\newblock Drone-aided delivery methods, challenge, and the future: A methodological review.
\newblock {\em Drones}, 7(3):191, 2023.

\bibitem{dutta2020application}
Gopal Dutta and Purba Goswami.
\newblock Application of drone in agriculture: A review.
\newblock {\em International Journal of Chemical Studies}, 8(5):181--187, 2020.

\bibitem{harrigan2024thermography}
Scott Harrigan and Harkin Aerial.
\newblock Thermography using unmanned aerial vehicles.
\newblock In {\em 3D Imaging of the Environment}, pages 175--189. CRC Press, 2024.

\bibitem{xu20213d}
Cheng Xu, Chanjuan Yin, Daqing Huang, Wei Han, and Dongzhen Wang.
\newblock 3d target localization based on multi--unmanned aerial vehicle cooperation.
\newblock {\em Measurement and Control}, 54(5-6):895--907, 2021.

\bibitem{men2021cooperative}
Tong Men, Daqian Liu, Xiaomin Zhu, Bowen Fei, Zhenliang Xiao, and Wenhua Xiao.
\newblock Cooperative localization method of uavs for a persistent surveillance task.
\newblock In {\em 2021 7th International Conference on Big Data and Information Analytics (BigDIA)}, pages 456--463. IEEE, 2021.

\bibitem{philion2020lift}
Jonah Philion and Sanja Fidler.
\newblock Lift, splat, shoot: Encoding images from arbitrary camera rigs by implicitly unprojecting to 3d.
\newblock In {\em Computer Vision--ECCV 2020: 16th European Conference, Glasgow, UK, August 23--28, 2020, Proceedings, Part XIV 16}, pages 194--210. Springer, 2020.

\bibitem{dosovitskiy2017carla}
Alexey Dosovitskiy, German Ros, Felipe Codevilla, Antonio Lopez, and Vladlen Koltun.
\newblock Carla: An open urban driving simulator.
\newblock In {\em Conference on robot learning}, pages 1--16. PMLR, 2017.

\bibitem{yan2023air}
Zhiyuan Yan, Peijin Wang, Feng Xu, Xian Sun, and Wenhui Diao.
\newblock Air-pv: A benchmark dataset for photovoltaic panel extraction in optical remote sensing imagery, 2023.

\bibitem{sun2023single}
Xian Sun, Yu~Tian, Wanxuan Lu, Peijin Wang, Ruigang Niu, Hongfeng Yu, and Kun Fu.
\newblock From single-to multi-modal remote sensing imagery interpretation: A survey and taxonomy.
\newblock {\em Science China Information Sciences}, 66(4):140301, 2023.

\bibitem{hu2023aerial}
Yue Hu, Shaoheng Fang, Weidi Xie, and Siheng Chen.
\newblock Aerial monocular 3d object detection.
\newblock {\em IEEE Robotics and Automation Letters}, 8(4):1959--1966, 2023.

\bibitem{huang2021bevdet}
Junjie Huang, Guan Huang, Zheng Zhu, Yun Ye, and Dalong Du.
\newblock Bevdet: High-performance multi-camera 3d object detection in bird-eye-view.
\newblock {\em arXiv preprint arXiv:2112.11790}, 2021.

\bibitem{huang2022bevdet4d}
Junjie Huang and Guan Huang.
\newblock Bevdet4d: Exploit temporal cues in multi-camera 3d object detection.
\newblock {\em arXiv preprint arXiv:2203.17054}, 2022.

\bibitem{li2023bevdepth}
Yinhao Li, Zheng Ge, Guanyi Yu, Jinrong Yang, Zengran Wang, Yukang Shi, Jianjian Sun, and Zeming Li.
\newblock Bevdepth: Acquisition of reliable depth for multi-view 3d object detection.
\newblock In {\em Proceedings of the AAAI Conference on Artificial Intelligence}, volume~37, pages 1477--1485, 2023.

\bibitem{zhou2023matrixvt}
Hongyu Zhou, Zheng Ge, Zeming Li, and Xiangyu Zhang.
\newblock Matrixvt: Efficient multi-camera to bev transformation for 3d perception.
\newblock In {\em Proceedings of the IEEE/CVF International Conference on Computer Vision}, pages 8548--8557, 2023.

\bibitem{wang2022detr3d}
Yue Wang, Vitor~Campagnolo Guizilini, Tianyuan Zhang, Yilun Wang, Hang Zhao, and Justin Solomon.
\newblock Detr3d: 3d object detection from multi-view images via 3d-to-2d queries.
\newblock In {\em Conference on Robot Learning}, pages 180--191. PMLR, 2022.

\bibitem{liu2022petr}
Yingfei Liu, Tiancai Wang, Xiangyu Zhang, and Jian Sun.
\newblock Petr: Position embedding transformation for multi-view 3d object detection.
\newblock In {\em European Conference on Computer Vision}, pages 531--548. Springer, 2022.

\bibitem{li2022bevformer}
Zhiqi Li, Wenhai Wang, Hongyang Li, Enze Xie, Chonghao Sima, Tong Lu, Yu~Qiao, and Jifeng Dai.
\newblock Bevformer: Learning bird’s-eye-view representation from multi-camera images via spatiotemporal transformers.
\newblock In {\em European conference on computer vision}, pages 1--18. Springer, 2022.

\bibitem{jiang2023far3d}
Xiaohui Jiang, Shuailin Li, Yingfei Liu, Shihao Wang, Fan Jia, Tiancai Wang, Lijin Han, and Xiangyu Zhang.
\newblock Far3d: Expanding the horizon for surround-view 3d object detection.
\newblock {\em arXiv preprint arXiv:2308.09616}, 2023.

\bibitem{wang2023exploring}
Shihao Wang, Yingfei Liu, Tiancai Wang, Ying Li, and Xiangyu Zhang.
\newblock Exploring object-centric temporal modeling for efficient multi-view 3d object detection.
\newblock In {\em Proceedings of the IEEE/CVF International Conference on Computer Vision}, pages 3621--3631, 2023.

\bibitem{liu2023sparsebev}
Haisong Liu, Yao Teng, Tao Lu, Haiguang Wang, and Limin Wang.
\newblock Sparsebev: High-performance sparse 3d object detection from multi-camera videos.
\newblock In {\em Proceedings of the IEEE/CVF International Conference on Computer Vision}, pages 18580--18590, 2023.

\bibitem{reading2021categorical}
Cody Reading, Ali Harakeh, Julia Chae, and Steven~L Waslander.
\newblock Categorical depth distribution network for monocular 3d object detection.
\newblock In {\em Proceedings of the IEEE/CVF Conference on Computer Vision and Pattern Recognition}, pages 8555--8564, 2021.

\bibitem{hartley2003multiple}
Richard Hartley and Andrew Zisserman.
\newblock {\em Multiple view geometry in computer vision}.
\newblock Cambridge university press, 2003.

\bibitem{shah2018airsim}
Shital Shah, Debadeepta Dey, Chris Lovett, and Ashish Kapoor.
\newblock Airsim: High-fidelity visual and physical simulation for autonomous vehicles.
\newblock In {\em Field and Service Robotics: Results of the 11th International Conference}, pages 621--635. Springer, 2018.

\bibitem{yin2021center}
Tianwei Yin, Xingyi Zhou, and Philipp Krahenbuhl.
\newblock Center-based 3d object detection and tracking.
\newblock In {\em Proceedings of the IEEE/CVF conference on computer vision and pattern recognition}, pages 11784--11793, 2021.

\bibitem{loshchilov2017decoupled}
Ilya Loshchilov and Frank Hutter.
\newblock Decoupled weight decay regularization.
\newblock {\em arXiv preprint arXiv:1711.05101}, 2017.

\end{thebibliography}

\newpage
\appendix

\section{Supplemental Material}


\subsection{AeroCollab3D Dataset Details}

\begin{figure}[h]
  \centering
  \includegraphics[width=\linewidth]{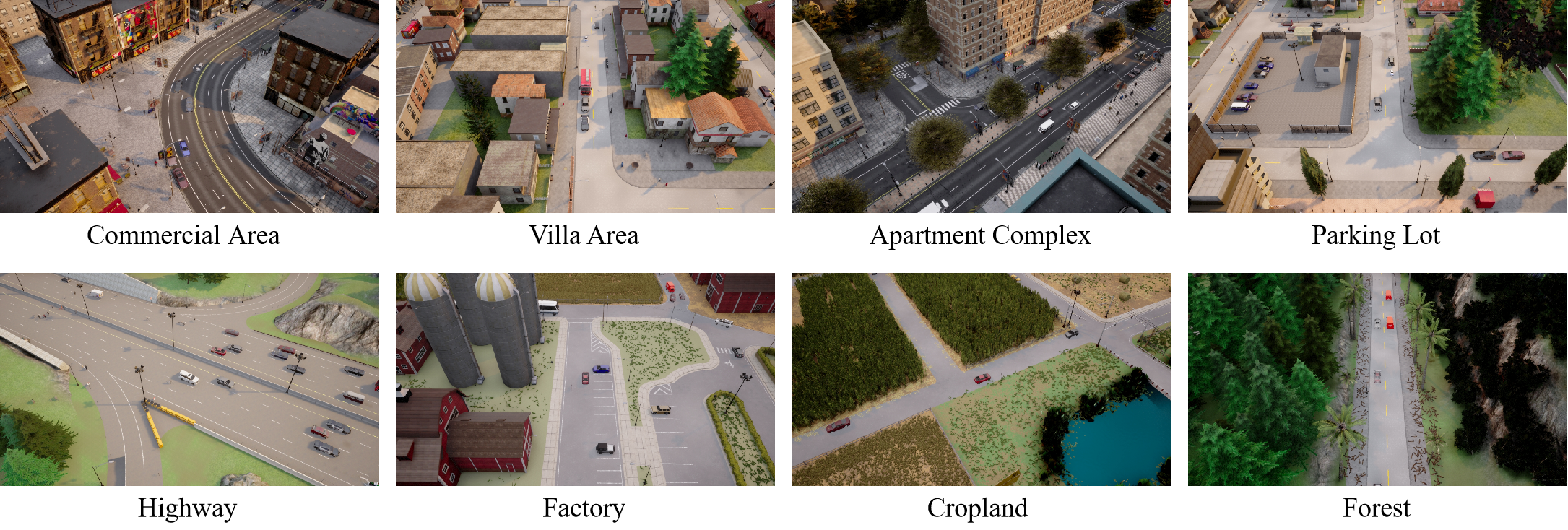}
  \caption{Visualization of Diverse Scenes in AeroCollab3D dataset.}
  \label{fig:aero_scene}
\end{figure}

{\bf Map creation}
The simulated scenes, including road layout, architectural facilities, green landscape, static objects, and traffic flow, are created based on CARLA simulation platform. We take eight open-source maps (Town 1 to Town 7 and Town 10) as the foundational road layout. The scenarios are diverse and complex, ranging from open fields to urban landscapes with observation occlusions, as shown in Fig.\ref{fig:aero_scene}, which provides a reliable foundation for verifying the effectiveness and generalization of multi-UAV collaborative 3D object detection.

{\bf Traffic Flow Creation}
Moving vehicles and people in the scene are controlled via CARLA, with hundreds of objects spawned in each scene using the official script provided by CARLA. The map’s road layout determines each object’s original location and motion trajectory.

{\bf Sensor Setup}
We set 6 collaborative UAVs flying at the height of 50m, each of which equips with only 1 camera at a pitch degree of -45$^\circ$ to simulate the real situation. The image resolution is 1600 $\times$ 900, and the BEV detection range is 110m × 110m. We set different world coordinates for each scene and different ego coordinates for each sample in the same scene. Each camera’s translation (x, y, z) and rotation (w, x, y, z in quaternion) are recorded in ego coordinates. And transformations from each ego coordinate to the world coordinate are also annotated.

\begin{figure}[h]
  \centering
  \includegraphics[width=\linewidth]{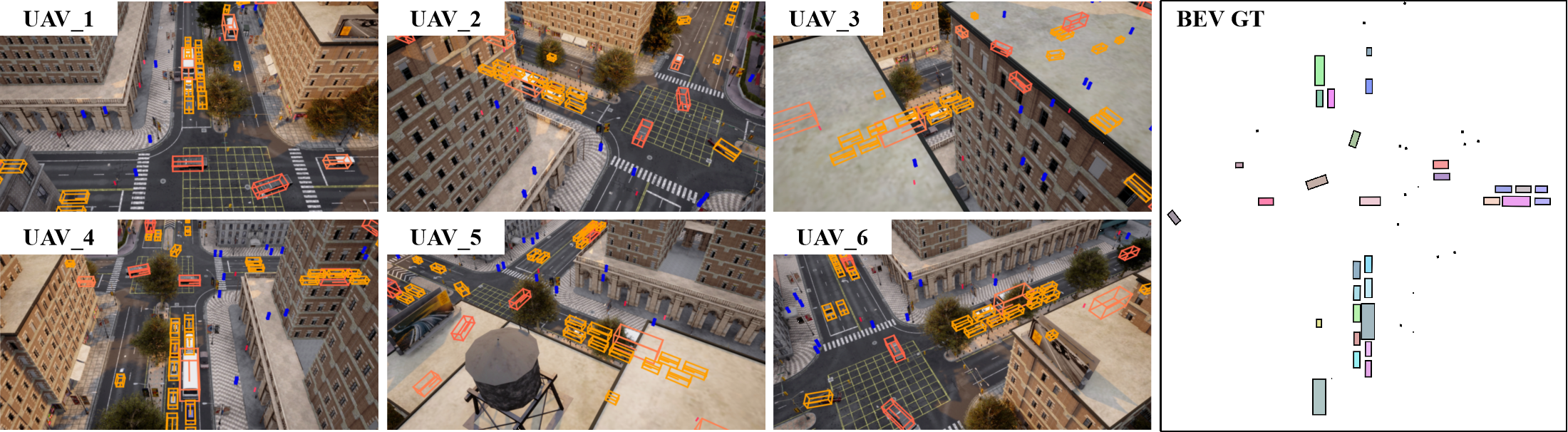}
  \caption{Visualization of Annotations in AeroCollab3D dataset.}
  \label{fig:aero_GT}
\end{figure}

{\bf Data Collection}
Our proposed dataset is collected by the CARLA simulation platform under the MIT license. We utilize CARLA to create complex simulation scenes and traffic flow. For each map, we set 20 scenes by randomly initializing the positions. For each scene, we set 20 samples which are collected at a frequency of 2 Hz. We synchronously collect images from 6 UAVs, resulting in 6 images per sample. A total of 19.2K images have been collected to support our experiments. Our ground truth labels for collaborative perception are derived from 3D bounding boxes of observed targets. Thus, we take advantage of lidar sensors to collect these 3D bounding boxes, which include location (x, y, z), rotation (represented with quaternion), and dimensions (length, width and height), amounting to nearly 21.8K 3D bounding boxes. For the same object, we marked its occurrence in different sample, so that our dataset can be used for multiple tasks such as detection, tracking, and prediction. Fig.\ref{fig:aero_GT} shows the annotations of AeroCollab3D dataset.

{\bf Data Usage}
We have randomly split the samples into training and validation sets, yielding 15.36k images for training and 3.84k for validation. The dataset is structured similarly to the widely used autonomous driving dataset, nuScenes, so that it can be directly used with the well-established nuScenes-devkit.

\subsection{Ground Depth Prior Derivation in GFM Module}
The intrinsic parameters of the camera are denoted by $\mathbf{K}\in \mathbb{R}^{3\times 3}$, while $\mathbf{R}\in \mathbb{R}^{3\times 3}$ and $\mathbf{T}\in \mathbb{R}^{3\times 1}$ respectively represent the rotation and translation matrices of the extrinsic parameters, which are known from the dataset. Based on the camera model, the transformation between a point $B=(x_{w}, y_{w}, z_{w})$ on the ground in the world coordinate system and its projection $P=(u, v)$ in the pixel coordinate system can be described as:

\begin{align}
l_{B}\begin{bmatrix}u\\v\\1\end{bmatrix}=\mathbf{K}(\mathbf{R}\begin{bmatrix}x_{w}\\y_{w}\\z_{w}\end{bmatrix}+\mathbf{T})
\end{align}

We denote $\mathbf{R}^{-1}\mathbf{K^{-1}}$ as the matrix $\mathbf{M}= (m_{ij} ) \in \mathbb{R}^{3\times 3}$, and $\mathbf{R}^{-1}(-\mathbf{T})$ as the vector $\mathbf{N}= (n_{i} ) \in \mathbb{R}^{3\times 1}$. Then we get the following formula:

\begin{align}
\begin{bmatrix}x_{w}\\y_{w}\\z_{w}\end{bmatrix}=\mathbf{R}^{-1}(\mathbf{K}^{-1}\begin{bmatrix}u\\v\\1\end{bmatrix}l_{B}-\mathbf{T})=
\begin{bmatrix}
 m_{11} & m_{12} & m_{13}\\
 m_{21} & m_{22} & m_{23}\\
 m_{31} & m_{32} & m_{33}
\end{bmatrix}\begin{bmatrix}
u \\
v \\
1
\end{bmatrix}l_B+
\begin{bmatrix}
n_1 \\
n_2 \\
n_3
\end{bmatrix}
\end{align}

Then the ground depth prior $l_B$ is

\begin{align}
l_{B} = \frac{z_{w}-n_{3}}{m_{31}u+m_{32}v+m_{33}} = \frac{z_{w}-[\mathbf{R}^{-1}(-\mathbf{T})]_{3}}{[\mathbf{R}^{-1}\mathbf{K^{-1}}]_{31}u+[\mathbf{R}^{-1}\mathbf{K^{-1}}]_{32}v+[\mathbf{R}^{-1}\mathbf{K^{-1}}]_{33}} 
\end{align}

\subsection{More Experiments Content}

\textbf{Implementation Details supplement}
We follow the setup from the baseline BEVDet4D. Initially, the raw images, with a resolution of 900 $\times$ 1600 pixels in AeroCollab3D and 450 $\times$ 900 in CoPerception-UAVs, are scaled and cropped to a size of 256 $\times$ 480.
As to view transformation, the $l_{BA}$ estimation range is configured from 0 to 10 meters, discretized into 10 intervals. Subsequently, for BEV representations, the spatial ranges for the x, y, and z axes are set to [-51.2, 51.2], [-51.2, 51.2], and [-5, 3] meters, respectively. We evaluate the model performance across various perceptual scopes: a 102.4 m $\times$ 102.4 m area with 0.8 m resolution. Moreover, we choose 200 homologous point pairs per batch in HPL.

\begin{figure}[!h]
  \centering
  \includegraphics[width=\linewidth]{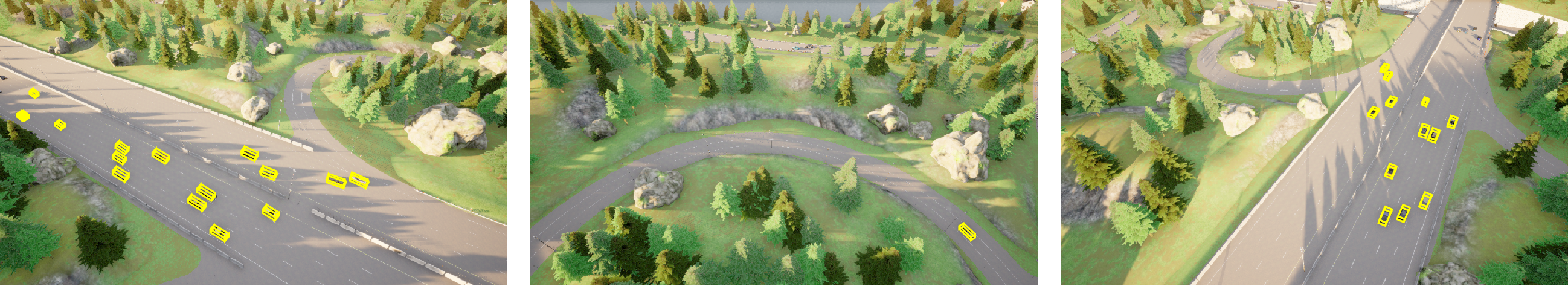}
  \caption{Visualization Results of CoPerception-UAVs dataset.}
  \label{fig:coUAV}
\end{figure}

\textbf{Visualization Results on CoPerception-UAVs Dataset.}
As depicted in Fig. \ref{fig:coUAV}, our method achieves good detection performance on the CoPerception-UAVs dataset. However, there are missed detections in the images due to the small object size.

\begin{figure}[!h]
  \centering
  \includegraphics[width=\linewidth]{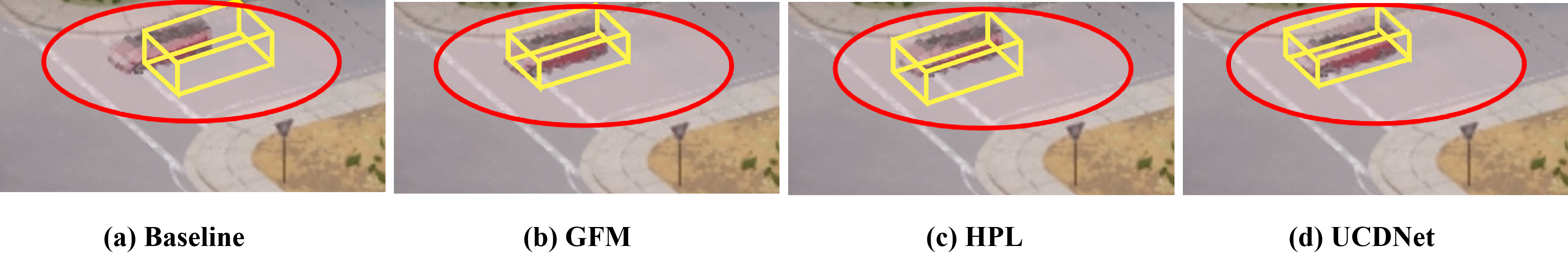}
  \caption{Visualization Results on Module Ablation.}
  \label{fig:moduleabla}
\end{figure}

\textbf{Visualization Results on Module Ablation.}
Fig. \ref{fig:moduleabla} shows the detection results after adding GFM and HPL respectively, where the two modules all enhance the 3D object localization. 

\end{document}